\documentclass[runningheads]{llncs}
\usepackage[T1]{fontenc}
\usepackage{color}  
\usepackage[table]{xcolor}
\usepackage{graphicx}
\usepackage{subfiles}
\usepackage{booktabs}
\usepackage{multirow}
\usepackage{caption}
\usepackage{latexsym}
\usepackage {marvosym}
\usepackage{amsmath}
\usepackage{amssymb}
\usepackage{arydshln}

\usepackage[colorlinks,linkcolor=blue]{hyperref}

\usepackage{amssymb}
\usepackage{latexsym}
\usepackage{makecell}
\usepackage{url}
\usepackage{xcolor}

\usepackage{comment}
\usepackage{hyperref}

\begin{document}
\title{EM-Net: Efficient Channel and Frequency Learning with Mamba for 3D Medical Image Segmentation}

\titlerunning{EM-Net}

\author{Ao Chang\inst{1,2,3}\thanks{Ao Chang and Jiajun Zeng contribute equally to this work.}, Jiajun Zeng \inst{1,2,3\star}, Ruobing Huang\inst{1,2,3} \textsuperscript{(\Letter)}, Dong Ni\inst{1,2,3} }

\authorrunning{Chang et al.}

\institute{\textsuperscript{$1$}National-Regional Key Technology Engineering Laboratory for Medical Ultrasound, School of Biomedical Engineering, Medical School, Shenzhen University, China\\
\email{ruobing.huang@szu.edu.cn} \\
\textsuperscript{$2$}Medical Ultrasound Image Computing (MUSIC) Lab, Shenzhen University, China\\
\textsuperscript{$3$}Marshall Laboratory of Biomedical Engineering, Shenzhen University, China}
\maketitle

\begin{abstract}

Convolutional neural networks have primarily led 3D medical image segmentation but may be limited by small receptive fields. Transformer models excel in capturing global relationships through self-attention but are challenged by high computational costs at high resolutions. Recently, Mamba, a state space model, has emerged as an effective approach for sequential modeling.
Inspired by its success, we introduce a novel Mamba-based 3D medical image segmentation model called EM-Net. It not only efficiently captures attentive interaction between regions by integrating and selecting channels, but also effectively utilizes frequency domain to harmonize the learning of features across varying scales, while accelerating training speed. Comprehensive experiments on two challenging multi-organ datasets with other state-of-the-art (SOTA) algorithms show that our method exhibits better segmentation accuracy while requiring nearly half the parameter size of SOTA models and 2x faster training speed.
Our code is publicly available at \url{https://github.com/zang0902/EM-Net}.
\end{abstract}

\section{Introduction}

3D medical image segmentation plays a crucial role in medical research, clinical decision-making, and surgical simulations as it enables precise visualization and analysis of volumetric data. In the past decade, convolutional neural networks (CNNs)~\cite{inproceedings} and transformers~\cite{attention} have emerged as the two mainstream approaches to address this problem.
Earlier methods mostly relied on the former, and have demonstrated impressive capabilities in extracting hierarchical image features (e.g.,  3D U-Net~\cite{3DUnet}, SegResNet~\cite{segresnet}, nn-UNet~\cite{nnU-Net}, and 3D UX-Net~\cite{3DUX-Net}). As dataset scales and hardware evolve, the transformer-based models ~\cite{dosovitskiy2020vit} models have garnered increased attention recently owing to their adeptness in capturing global information.
Many state-of-the-art (SOTA) methods like UNETR~\cite{hatamizadeh2022unetr}, 3D TransUNet~\cite{chen2023transunet3d}, and Swin UNETR~\cite{SwinUNETR} have explored ways to integrate transformers into CNNs to better solve 3D image segmentation. However, the self-attention mechanism in transformers, while effective for processing sequences, becomes less efficient when applied to 3D images as its computation is accompanied by quadratic complexity that escalates rapidly with the volume size.
Recently, state space models (SSMs)~\cite{NEURIPS2021_05546b0e,mamba} were introduced as effective building blocks for constructing deep networks to analyze long sequences of data~\cite{gu2022efficiently}. 
For example, Mamba~\cite{mamba} has demonstrated SOTA performance in processing language and genomics data with virtue of linear computational complexity and strong global awareness. 
Furthermore, the potential of Mamba in image analysis is evidenced by Vmamba~\cite{liu2024vmamba} and Vim~\cite{vim}, which analyze two-dimensional natural images.
Meanwhile, U-Mamba~\cite{ma2024umamba} has been optimized for medical imaging, further demonstrating its broad applicability and flexibility. 

In this work, we envision that the Mamba-based models, which excel in compressing and summarizing features, may also serve as an effective solution to address 3D medical segmentation. However, learning the spatial relationship between 3D visual patches is still non-trivial for sequential models and often solely relies on strong positional encoding. Moreover, medical segmentation often requires extensive memory consumption given restricted hardware support. To address these, we propose a novel efficient 3D medical image segmentation framework, named EM-Net. Our contribution is three-fold:
\textbf{First}, we propose a novel Mamba-based architecture for 3D medical segmentation. It is equipped with channel squeeze-reinforce Mamba (CSRM) blocks that learn to attend to selective regions.
\textbf{Second}, we propose the EFL (efficient frequency-domain learning) layer that enables learnable frequency weighting to harmonize the learning of features across varying scales.
\textbf{Third}, a Mamba-infused decoder to further improve segmentation performance while suppressing memory costs. 
Experimental results demonstrate that the EM-Net performs well in two challenging 3D multi-organ segmentation datasets while requiring nearly half the parameter size of SOTA models and 2x faster training speed.

\begin{figure}[t]
\centering
\includegraphics[width=1\textwidth]{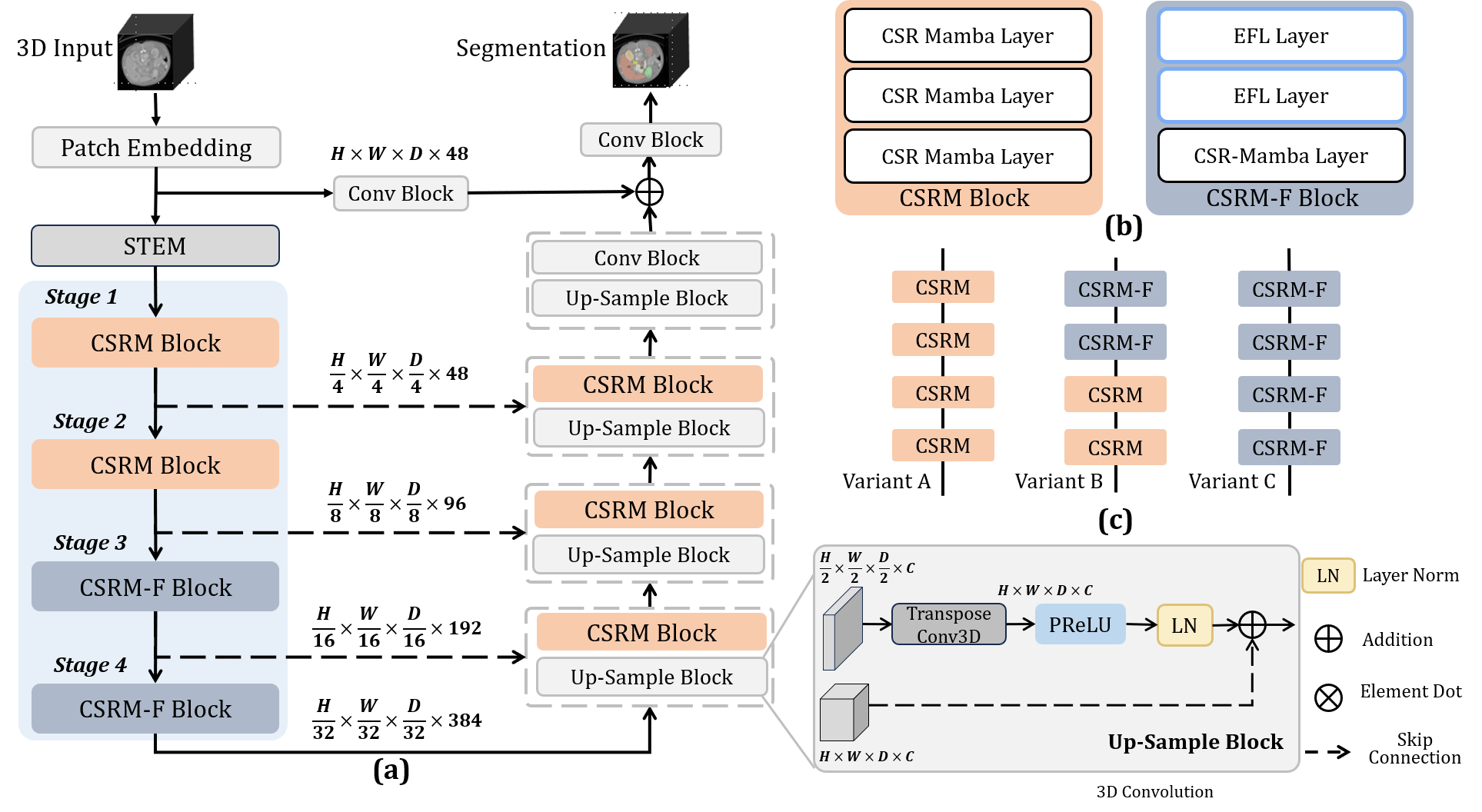}
\caption{(a) Overview of our proposed EM-Net framework. (b) shows CSRM and CSRM-F block. (c) shows our three variants of encoder design.} \label{fig1}
\end{figure}
\section{Methodology}

Fig.~\ref{fig1} illustrates the overall framework of the proposed EM-Net. In this section, we first explain the overall encoder-decoder architecture, and then elaborate on the design of each key component. 
Specifically, we first divide each 3D input $x \in \mathbb{R}^{H\times W\times D}$ into $N$ non-overlapping patches $x_u \in \mathbb{R}^{P_1\times P_2\times P_3}$, where $P_1, P_2, P_3$ represents the scale of each dimension, and $N = \frac{H}{P_1}\times \frac{W}{P_2} \times \frac{D}{P_3}$ denotes the length of the sequence. 
Then, the STEM layer downsamples the embedding using convolution with a kernel size of $4\times4\times4$ and stride of $4\times4\times4$, acquiring feature with the size of $\frac{H}{4}\times \frac{W}{4} \times \frac{D}{4}$.
The output is then passed to two consecutive CSRM blocks and two CSRM-F (Frequency) blocks to extract multi-scale features. The CSRM-F block is equipped with EFL layers to further improve the computation efficiency (see Fig.~\ref{fig1} (b)). Note that besides the design we have chosen, there are also variants available with different arrangements and selections of blocks (e.g., Fig.~\ref{fig1} (c)) which we explored later in ablation studies. 
Similar to the encoder, the decoder also comprises four stages, with each decoder stage including a double-resolution upsampling layer using deconvolution followed by a CSRM block. Finally, the output of the decoder is transmitted through a convolution block to predict the segmentation mask. Next, we explain the design of each key component in detail.

\begin{figure}
\centering
\includegraphics[width=1.\textwidth]{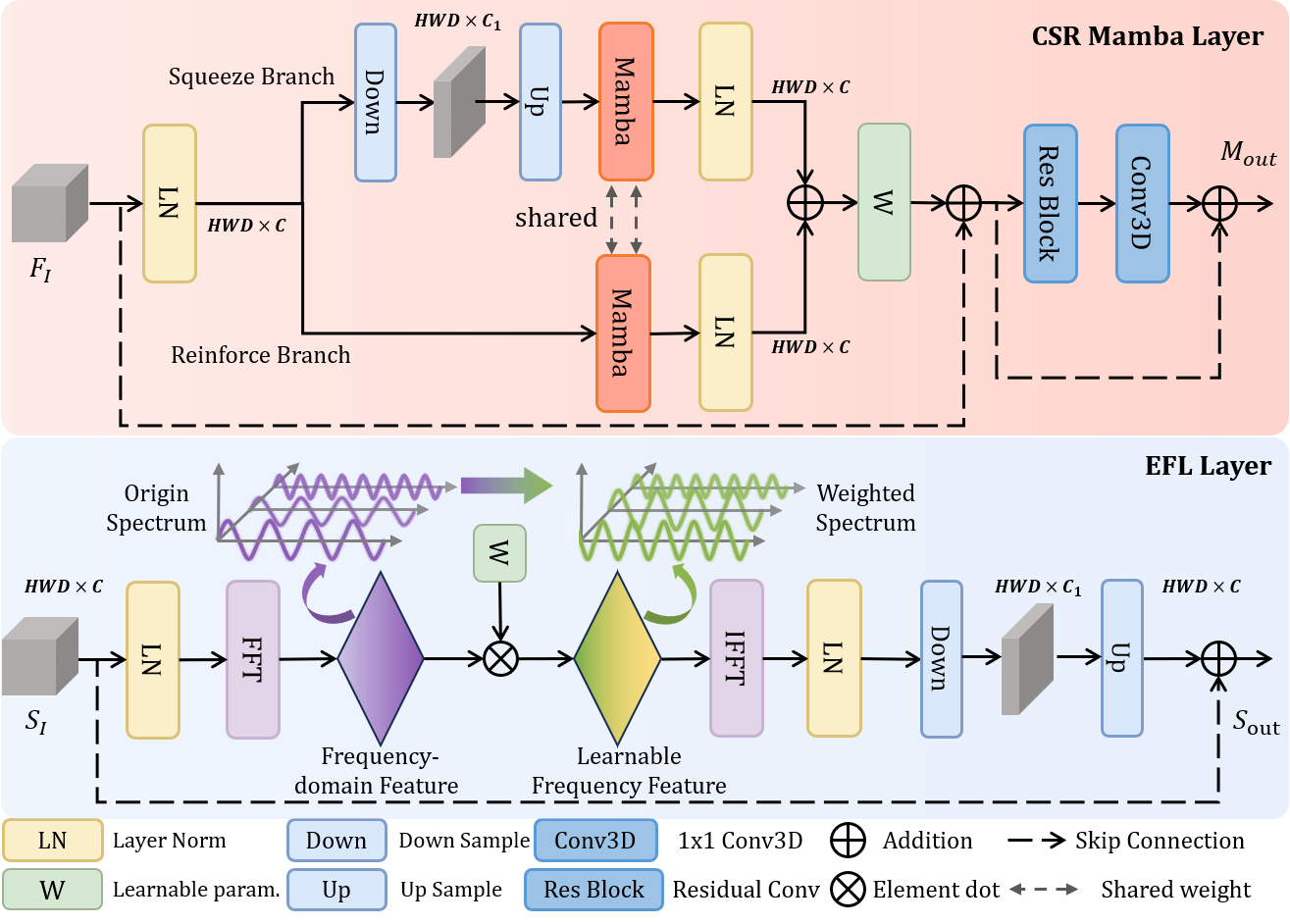}
\caption{Overview of our proposed method. } 
\label{fig:method}
\end{figure}

\subsection{Encoder in EM-Net}
\textbf{CSR Mamba layer and CSRM block.}
An enduring challenge in utilizing sequential models for image analysis is the difficulty in modeling spatial relationships. Positional encoding may partly address this issue while it may still be limited in representing complex spatial dependencies among 3D patches. To mitigate this, we propose to seek assistance from the channel dimension. Diverse channels may emphasize different image regions each with various importance to the segmentation task.
The proposed CSR Mamba layer attempts to diminish less relevant channels and calibrate itself to capture relevant regional patterns of each target. In specific, it includes two branches: \textit{squeeze} and \textit{reinforce}.
The squeeze branch is designated for channel selection, compressing channel dimension to filter features and then expanding to retain the essential ones. Meanwhile, the reinforce branch is used for adaptive calibration of overlooked features in the squeeze branch.
These two branches multiplex the same Mamba module to reduce computation while acquiring mutual information to ensure accuracy.
As illustrated in Fig.~\ref{fig:method} (upper), the input feature map $F_{I}$ is fed into the squeeze and reinforce branches, respectively,  resulting in $M_{s} = {Mamba}_{shared}(Up(Down(F_{I})))$ and 
$M_{e} = {Mamba}_{shared}(F_{I})$. Their results are merged following: $M = F_{I} + \omega(M_{s}+M_{e}))$, where $\omega$ is a learnable parameter.
The final output can be obtained as:
\begin{align}
    M_{out} = M + Conv(Resblock(M)),
    \label{eq4}
\end{align}
where $Resblock$ denotes Residual Block~\cite{nnU-Net}.

\textbf{EFL layer and CSRM-F block.}
Another commonly encountered issue in segmentation is the trade-off between global and local features. Different from classification, accurate segmentation requires both global understandings (corresponds to signals with low frequencies), and local textural information (corresponds to signals with high frequencies)~\cite{wang2024fremim}.  
As the frequency domain (FD) comprehensively represents both local and global signals~\cite{wang2024fremim}, we propose using the fast fourier transform (FFT) to build an EFL layer with log-linear computational complexity to achieve a balance between the two.
It can be regarded as a trainable frequency filter that automatically learns to highlight signals with diverse frequencies that are related to segmentation while suppressing the rest. Note that such operations could also help to relieve the computation burden. 

The EFL layer employs a spectral gated network, which consists of an FFT layer, followed by a weighted gating layer, and then an Inverse FFT (IFFT) layer (see Fig.~\ref{fig:method} lower). It converts the physical space into spectral space using FFT. We utilize learnable weight parameters to automatically learn the weight of each frequency component, thereby capturing both the rough and detailed spatial features. Note that these weights are specific to each FFT layer to suit each specific stage.
IFFT transforms the spectral space back to the spatial space. 
Layer normalization and multi-layer perceptron are used to mix channels.

Mathematically, given a feature map $S_{I}\in \mathbb{R}^{H\times W\times D\times C}$, we first perform a 3D FFT along the spatial dimensions, which can be formulated as: 
\begin{equation}
   M = F[S_{I}] = \sum_{x=0}^{H-1} \sum_{y=0}^{W-1} \sum_{z=0}^{D-1} M(x, y, z) \cdot e^{-j2\pi(m_x \cdot x + m_y \cdot y + m_z \cdot z) / (HWD)},
\end{equation}
where $F[\cdot]$ denotes the 3D FFT. Then, we modulate the spectrum of M by multiplying a parameterized attention map $ A\in \mathbb{R}^{H\times W\times D\times C}$ to M:
\begin{equation}
    M' = A \otimes M,
\end{equation}
where $\otimes$ denotes the element-wise product. Finally, we reverse $M'$ back to the spatial domain by adopting inverse FFT to obtain $m' = F^{-1}[M']$. The output of the CSRM-F block can be defined as $S_{out} = S_I + Up(Down(m'))$.

\subsection{Decoder in EM-Net}
Most 3D medical image segmentation models simply use an upsampling block and a convolution block as their decoder. Despite the popularity of this up-conv design, we argue that it is also essential to equip the decoder with strong feature selection and integration ability for image segmentation tasks. In fact, our preliminary experiments demonstrated that over-elaborated architecture for the decoder may result in a decrease in the computational speed while hampering the model performance. Thus, we also insert the proposed CSRM block in the decoding stage to improve the model performance while minimizing additional computational and memory consumption. Therefore, the decoder consists of four serial blocks, with an upsampling module preceding each block, used to increase the resolution. Following that, we place the CSRM block at the deepest three stages, and the convolution block at the first stage. The number of channels is halved after each stage. Finally, the output of the last decoder is fused with the feature map to recover spatial information and enhance feature representation.
 
\section{Experiment}
\subsection{Datasets and implementation}

\textbf{Datasets.}
We validate our proposed framework on two datasets: \textbf{Synapse}~\cite{synapse_dataset}, and \textbf{BTCV} for multi-organ segmentation in CT~\cite{synapse_dataset}.
The Synapse dataset comprises abdominal CT scans from 30 subjects, encompassing 8 organs: spleen, right kidney, left kidney, gallbladder, liver, stomach, aorta, and pancreas. 
The BTCV~\cite{synapse_dataset} dataset consists of 13 organs, including 8 organs of Synapse, along with the esophagus, inferior vena cava, portal and splenic veins, right and left adrenal gland. Both datasets are split into 18 training and 12 validation sets. 

\textbf{Implementation Details.}
We implemented our framework using PyTorch using one NVIDIA A100 GPU with 80 GB of memory. 
The input resolutions for the Synapse and BTCV datasets were $128^3$ and $96^3$, respectively. 
The loss function was DiceCE Loss. We ran 500 epochs for all datasets and used SGD optimizer, initial learning rate of 0.01, a decay of 1e-5. 
We applied overlapped sliding window inference~\cite{nnU-Net} with an overlap ratio set to 0.5.

\textbf{Evaluation Metrics.} Performance was evaluated using the dice similarity coefficient (\%) (DSC), and hausdorff distance (HD)  for all the experiments. 

\subsection{Comparison Methods}
We compared the EM-Net with ten SOTA methods, including CNN-based methods (3D U-Net~\cite{3DUnet}, SegresNet~\cite{segresnet}, nn-UNet~\cite{nnU-Net} and 3D UX-Net~\cite{3DUX-Net}), diffusion-based method Diff-UNet~\cite{xing2023diff}, transformer-based methods (VT-UNet~\cite{vt-unet}, UNETR~\cite{hatamizadeh2022unetr}, Swin UNETR~\cite{SwinUNETR}, 3D TransUNet~\cite{chen2023transunet3d}), and Mamba-based method U-Mamba Bot~\cite{ma2024umamba}, denoted as U-Mamba.
For a fair comparison, we used publicly available codes for all methods. We reported segmentation performance solely on individual model accuracy, without using pre-training, model ensembling, or extra data. The best results were highlighted in bold.

In Table~\ref{tab:tab1}, it can be observed that our EM-Net achieved an average DSC of 83.95\%, higher than the current SOTA segmentation algorithms, i.e., U-Mamba (82.83\%) and Swin UNETR (83.06\%). 
Our method yielded the highest DSC  in segmenting a variety of organs, e.g., spleen (Spl), aorta (Aor), and right kidney (RKid), at 92.75\%, 91.24\%, and 86.88\%, respectively. 
Note that the EM-net also excelled in recognizing difficult objects such as small organs, with a DSC of 71.5\% in segmenting the gallbladder (Gal) which is significantly higher than that of the U-Mamba (63.98). It is worth noting that EM-net achieved a notably higher DSC of 69.42\% in segmenting the pancreas (Pan), which was significantly superior to the DSC values obtained by other methods such as U-Mamba and Diff-UNet, which achieved 68.27\% and 67.02\%, respectively.
In terms of the HD metrics, EM-Net also performed optimally and reported a score of 21.16, 46\% less than that of the 3D U-Net. 

In Table~\ref{tab:tab2}, EM-Net exhibited supreme performance, attaining the best average DSC (78.97\%) and HD (25.24).
In the challenging organs, the esophagus (Eso) and adrenal gland (AG indicates the mean of the left and right adrenal gland), EM-Net achieved the best results (reaching 75.76\% and 64.91\%). These results are superior to those of U-Mamba and nn-UNet, with U-Mamba achieving 73.37\% and 61.97\%, and nn-UNet achieving 76.85\% and 61.60\%. On the inferior vena cava (IVC), EM-Net also achieved a secondary best of 85.54\%, while still exceeding that of the 3D TransUNet by 8.19\%.
Our EM-Net showed robust performance across different organ sizes, from small ones like the pancreas and gallbladder to larger ones like the stomach and liver. The Wilcoxon signed-rank test confirmed its statistically significant superiority over other SOTA approaches (p < 0.05).
Note that we also validated the effectiveness of our EM-Net on additional publicly available datasets (i.e., Flare~\cite{FLARE22}), and the findings were consistent while details were omitted due to limited space.

\begin{table}[t!]
    \caption{Experimental results obtained from the \textbf{Synapse} dataset.}
    \centering
        \resizebox{\textwidth}{!}{
        \begin{tabular}{c|c|cccccccc|cc}
        \toprule  
            \multirow{2}{*}{Methods}& \multirow{2}{*}{Param.}& 
            \multirow{2}{*}{Spl} & \multirow{2}{*}{LKid} & \multirow{2}{*}{RKid} & \multirow{2}{*}{Gal} & \multirow{2}{*}{Liv} & \multirow{2}{*}{Sto} & \multirow{2}{*}{Aor} & \multirow{2}{*}{Pan} & \multicolumn{2}{c}{Average}\\
            \cline{11-12}
            & & & & & & & & & & DSC$\uparrow$&HD$\downarrow$  \\
            \hline
            3D U-Net\cite{3DUnet}& 4.81& 88.41  & 80.30  & 80.94  & 53.55  & 95.33  & 72.19  & 87.41  & 52.46  & 76.32  & 39.21\\
            SegresNet\cite{segresnet}& 4.70& 90.40  & 83.71  & 85.07  & 61.76  & 95.43  & 77.60  & 90.48  & 66.44  & 81.36  & 31.86  \\
            
            3D UX-Net\cite{3DUX-Net}& 53.01&91.18  & 85.64  & 86.16  & 60.04  & 96.09  & 73.96  & 90.84  & 59.91  & 80.48  & 35.40\\
            
            nn-UNet\cite{nnU-Net}&31.20&89.52 & 81.38 & 81.97 & 57.47 & 95.41 & 66.59 & 90.90 & 65.21 & 78.56 & 55.94\\
            Diff-UNet\cite{xing2023diff} & 38.38 &89.93&78.28&85.90&69.67&94.86&69.06&90.99&67.02&80.71&21.64\\
            VT-UNet\cite{vt-unet} &11.78&90.28  & 81.12  & 83.17  & 65.03  & 95.51  & 73.31  & 82.94  & 57.35  & 78.59  & 25.36\\

            UNETR\cite{hatamizadeh2022unetr}& 93.01  & 85.47  & 78.82  & 81.60  & 58.15  & 93.78  & 65.68  & 82.96  & 44.01  & 73.81  & 69.49\\
            Swin UNETR\cite{SwinUNETR} &62.19& 92.01  & 85.85  & 86.69  & 67.18  & 95.78  & 79.33  & 90.72  & 66.92  & 83.06  & 42.09 \\
            3D TransUNet\cite{chen2023transunet3d} & 42.98 &  91.49 & \textbf{85.75}  & 85.70  & 57.97  & 95.66  & 78.33  & 91.22 & 64.62  & 81.34 & 26.50\\

            U-Mamba\cite{ma2024umamba}&58.47& 92.33  & 85.83  & 86.31  & 63.98  & \textbf{96.22}  & \textbf{79.18}  & 90.55  & 68.27  & 82.83  & 27.17 \\
            
            \hline
            EM-Net&38.82& \textbf{92.75}  & 84.77  & \textbf{86.88}  & \textbf{71.50}  & 96.12 & 78.93  & \textbf{91.24}  & \textbf{69.42}  & \textbf{83.95}  &\textbf{21.16}  \\ 

        \bottomrule 
        \end{tabular}
    }
    \label{tab:tab1}
\end{table}

\begin{table}[t!]
    \caption{Experimental results obtained from the \textbf{BTCV} dataset.}
    \centering
        \resizebox{\textwidth}{!}{
        \begin{tabular}{c|cccccccccccc|cc}
        \toprule  
            \multirow{2}{*}{Methods}& 
            \multirow{2}{*}{Spl}&\multirow{2}{*}{RKid}&\multirow{2}{*}{LKid}& \multirow{2}{*}{Gall}&\multirow{2}{*}{Eso}&\multirow{2}{*}{Liv}&
            \multirow{2}{*}{Sto}&\multirow{2}{*}{Aor}&\multirow{2}{*}{IVC}&
            \multirow{2}{*}{Psv}&\multirow{2}{*}{Pan}&\multirow{2}{*}{AG}&
    
            \multicolumn{2}{c}{Average}\\
            \cline{14-15}
            &  & & & & & & & & & & & & DSC$\uparrow$&HD $\downarrow$  \\
            \midrule
            3D U-Net\cite{3DUnet}& 88.34  & 81.96  & 83.48  & 53.27  & 70.28  & 95.27  & 72.17  & 88.74  & 79.40  & 62.71  & 56.99  & 56.75  & 72.78  & 36.88\\
            SegresNet\cite{segresnet}&82.61  & 82.08  & 82.75  & 54.75  & 69.23  & 95.33  & 71.92  & 87.30  & 79.53  & 67.34  & 58.18  & 57.81  & 72.82  & 40.85\\
            3D UX-Net\cite{3DUX-Net}&88.16  & 85.09  & 86.07  & 60.04  & 75.07  & 95.32  & 70.54  & 90.56  & 85.76  & 67.20  & 63.03  & 61.00  & 76.07  & 35.78 \\
            nn-UNet\cite{nnU-Net}&88.45 & 84.25 & 85.01 & 58.85 & 76.85 & 95.77 & 69.59 & 88.90 & 83.09 & 68.19 & 69.88 & 61.60 & 76.31 & 43.81\\
            Diff-UNet\cite{xing2023diff}&90.46&79.49 &83.01& 58.21&46.97&93.22&
    70.89& 87.14& 80.14& 60.79& 53.74& 51.61&69.79&62.46\\
            VT-UNet\cite{vt-unet}&90.02  & 83.71  & 84.57  & 65.16  & 68.29  & 95.28  & 76.40  & 85.08  & 78.83  & 62.45  & 58.49 & 52.92  & 73.82  & 27.21  \\
            UNETR\cite{hatamizadeh2022unetr}&82.95  & 80.51  & 82.14  & 65.47  & 66.77  & 94.12  & 67.61  & 84.57  & 74.63  & 52.42  & 43.66  & 50.33  & 68.88  & 49.55 \\
            Swin UNETR\cite{SwinUNETR}&88.97  & 82.58  & 86.15  & 66.69  & 73.15  & 94.86  & 71.15  & 88.92  & 82.45  & 66.77  & 60.84  & 58.43  & 75.34  & 48.45  \\
            3D TransUNet\cite{chen2023transunet3d}&80.89  & 77.50  & 82.15  & 53.75  & 68.21  & 94.51  & 64.68  & 86.45  & 77.35  & 60.39  & 52.84  & 59.49 & 70.59  & 41.01 \\
            U-Mamba\cite{ma2024umamba}& \textbf{93.50}  & 81.12  & 85.81  & 66.32  & 73.37  & 95.86  & \textbf{80.65}  & 90.70  & \textbf{86.06}  & \textbf{70.96}  & 65.49  & 61.97  & 77.98  & 30.21\\
            
            \hline
            EM-Net&88.34  & \textbf{85.58}  & \textbf{86.33}  & \textbf{70.24}  & \textbf{75.76}  & \textbf{96.86}  & 79.31  & \textbf{91.16}  & 85.54  & 67.70  & \textbf{70.04}  & \textbf{64.91}  &\textbf{78.97}   & \textbf{25.24} \\       
        \bottomrule 
        \end{tabular}
    }
    \label{tab:tab2}
\end{table}

\begin{figure}[!t]
\centering
\includegraphics[width=1.\textwidth,height=.5\textwidth]{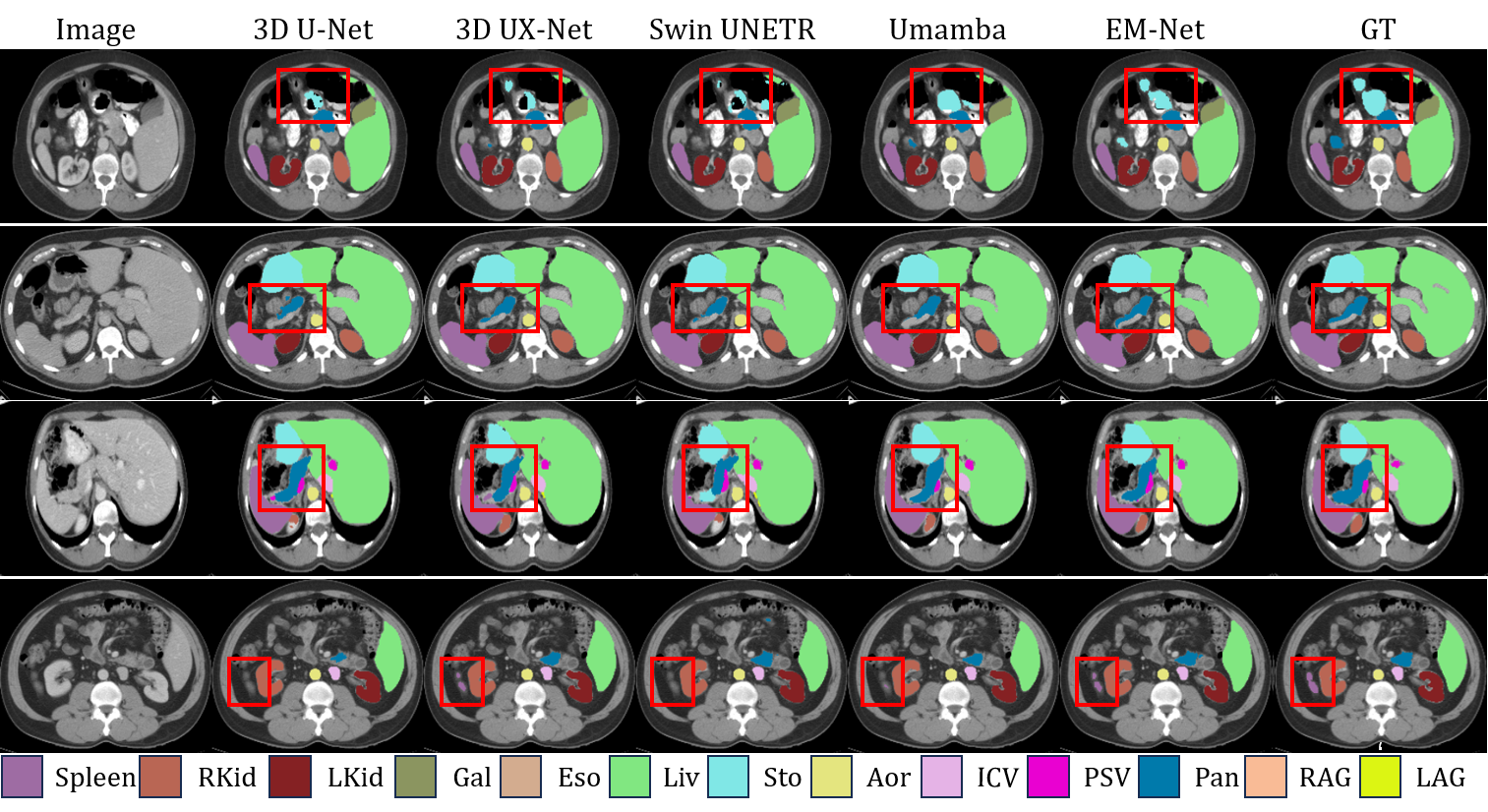}
\caption{Visual results. The red rectangles highlight challenging regions where the superiority of EM-Net is evident.} 
\label{fig:3d_result}
\end{figure}

Fig.~\ref{fig:3d_result} provides a qualitative visualization of the results, with the top two rows showing the Synapse dataset and the bottom ones corresponding to the BTCV dataset. The first row demonstrated that most competing methods face challenges in accurately segmenting the stomach. Contrarily, EM-Net exhibited superior performance in accurately delineating it, despite heterogeneous background texture and low contrast. Row 2 and 3 displayed 2 examples of the pancreas with large variations in size and orientation, while the EM-Net exhibited robust performance. The fourth row highlighted the precision of EM-Net in segmenting the spleen, whose prediction aligns well with the GT.

\begin{table}[t!]
    \caption{Comparison of model efficiency and ablation on \textbf{Synapse} dataset.}
    \centering
        \resizebox{1.\textwidth}{!}{
        \begin{tabular}{cccccccc}
        \toprule  
            Methods & Arch. & Param.(M) & Flops(G) & Memory(M) & TS$\uparrow$(Iter./s)&  \makebox[0.1\textwidth][c]{DSC$\uparrow$}  & \makebox[0.1\textwidth][c]{HD$\downarrow$}\\
            \midrule
            3D UX-Net\cite{3DUX-Net}&CNN&53.01 &2805.8 &8998& 3.96&80.48&35.40 \\
            Swin UNETR\cite{SwinUNETR}& Transformer & 62.19 &1505.3&19156&4.94&83.06&42.09\\
            U-Mamba\cite{ma2024umamba}& Mamba &58.47 &5620.0&12332&8.14&82.83&27.17\\
            \hline
            \rowcolor{gray!15}Variant A& Mamba & 62.09 &904.2&11694&10.45&84.07&20.98 \\
            \rowcolor{gray!15}Variant B& Mamba & 60.91&877.4&11512&11.25&83.39&30.35\\
            \rowcolor{gray!15}Variant C& Mamba & 38.23& 870.9&10870&11.14&82.13&30.41\\
            \rowcolor{gray!15}EM-Net& Mamba & 39.41& 897.7&11438&10.68&83.95&21.16\\
            
        \bottomrule 
        \end{tabular}
    }
    \label{tab:effiency}
\end{table}

\textbf{The high efficiency and ablation study of EM-Net.}
To verify the design of EM-Net and validate its efficiency and efficacy, we benchmarked it against several other SOTA methods and our own developed variants (-A, -B, and -C)(see Fig.~\ref{fig1}(b)).
Table~\ref{tab:effiency} showed that compared to 3D UX-Net with large kernel convolutions~\cite{3DUX-Net} and Swin UNETR~\cite{SwinUNETR}, our EM-Net achieved the highest DSC with only $39.41M$ parameters, which were significantly fewer than those of 3D UX-Net and Swin UNETR, and the memory usage was also half that of Swin UNETR. Compared to U-Mamba~\cite{ma2024umamba}, which also used the Mamba as its core component, our model not only had fewer parameters but also reached a higher training speed (TS) of 10.68, far ahead of that of the U-Mamba. 
Moreover, to validate the efficacy of the CSRM and CSRM-F module designs, we conducted extensive ablation experiments and compared our method with three variants, thereby providing further evidence (see Table~\ref{tab:effiency} rows 4-7).
Variant A, incorporating the full CSRM block, had more parameters and achieved the highest DSC and HD. On the other hand, variant C utilized the complete CSRM-F module, resulting in the smallest parameter count and faster training speed. However, due to its lack of Mamba's spatial feature extraction capability, it yielded a slightly lower DSC. Our proposed EM-Net effectively integrates the strengths of both, offering high segmentation accuracy while maintaining a low parameter count and enhanced training efficiency.

\section{Conclusion}
In this study, we introduce a novel efficient 3D medical image segmentation framework named EM-Net, which consists of the following components: 1) CSRM block for channel squeeze-reinforce Mamba that learns to attend to selective regions, 2) EFL layer for efficient frequency-domain learning, and 3) a Mamba-infused decoder to further improve segmentation performance while suppressing memory costs. Comprehensive experimental results reveal that EM-Net outperforms other SOTA methods across two datasets.

\begin{credits}
\subsubsection{\ackname}
This work was supported by National Natural Science Foundation of China (Nos.62171290, 62101343), Guangdong Basic and Applied Basic Research Foundation (No.2023A1515012960), Shenzhen-Hong Kong Joint Research Program (No.SGDX20201103095613036), Shenzhen Science and Technology Innovations Committee (No.20200812143441001).
\subsubsection{\discintname}
The authors have no competing interests to declare that are
relevant to the content of this article. 
\end{credits}

\bibliographystyle{splncs04}
\bibliography{main}

\end{document}